\documentclass{article}

\usepackage{microtype}
\usepackage{graphicx}
\usepackage{subcaption}
\usepackage{booktabs} 

\usepackage{hyperref}

\usepackage[preprint]{icml2026}

\usepackage{amsmath}
\usepackage{amssymb}
\usepackage{mathtools}
\usepackage{amsthm}
\usepackage{adjustbox}
\usepackage{multirow}
\usepackage{tabularx}
\usepackage{bbding}
\usepackage[table]{xcolor}
\usepackage{xltabular}
\usepackage{float}

\usepackage[capitalize,noabbrev]{cleveref}

\theoremstyle{plain}

\theoremstyle{definition}

\theoremstyle{remark}

\usepackage[textsize=tiny]{todonotes}

\icmltitlerunning{History-Guided Iterative Visual Reasoning with Self-Correction}

\begin{document}

\twocolumn[
  \icmltitle{History-Guided Iterative Visual Reasoning with Self-Correction}



  \icmlsetsymbol{equal}{*}

  \begin{icmlauthorlist}
    \icmlauthor{Xinglong Yang}{sch}
    \icmlauthor{Zhilin Peng}{sch}
    \icmlauthor{Zhanzhan Liu}{sch}
    \icmlauthor{Haochen Shi}{equal,sch}
    \icmlauthor{Sheng-Jun Huang}{equal,sch}

  \end{icmlauthorlist}

  \icmlaffiliation{sch}{Nanjing University of Aeronautics and Astronautics}

  \icmlcorrespondingauthor{Haochen Shi}{hshi03@nuaa.edu.cn}
  \icmlcorrespondingauthor{Sheng-Jun Huang}{huangsj@nuaa.edu.cn}

  \icmlkeywords{Machine Learning, ICML}

  \vskip 0.3in
]



\printAffiliationsAndNotice{\icmlEqualContribution}  

\begin{abstract}

Self-consistency methods are the core technique for improving the reasoning reliability of multimodal large language models (MLLMs). By generating multiple reasoning results through repeated sampling and selecting the best answer via voting, they play an important role in cross-modal tasks. However, most existing self-consistency methods are limited to a fixed ``repeated sampling and voting'' paradigm and do not reuse historical reasoning information. As a result, models struggle to actively correct visual understanding errors and dynamically adjust their reasoning during iteration. Inspired by the human reasoning behavior of repeated verification and dynamic error correction, we propose the H-GIVR framework. During iterative reasoning, the MLLM observes the image multiple times and uses previously generated answers as references for subsequent steps, enabling dynamic correction of errors and improving answer accuracy. We conduct comprehensive experiments on five datasets and three models. The results show that the H-GIVR framework can significantly improve cross-modal reasoning accuracy while maintaining low computational cost. For instance, using \texttt{Llama3.2-vision:11b} on the ScienceQA dataset, the model requires an average of 2.57 responses per question to achieve an accuracy of 78.90\%, representing a 107\% improvement over the baseline.
\end{abstract}

\section{Introduction}
In recent years, multimodal large language models have made significant progress in vision–language reasoning tasks, showing early abilities in multi-step reasoning and cross-modal alignment~\cite{li2022blip,alayrac2022flamingo,liu2023visual}. However, increasing model size to improve performance leads to an exponential growth in training cost, which indicates that model scaling alone is not a sustainable path~\cite{kaplan2020scaling,hoffmann2022training}. Therefore, prompt engineering has become one of the key research directions for improving the reasoning performance of large language  models.

\begin{figure*}[!t]
    \centering
    \setlength{\tabcolsep}{1pt}
    \includegraphics[width=1\textwidth]{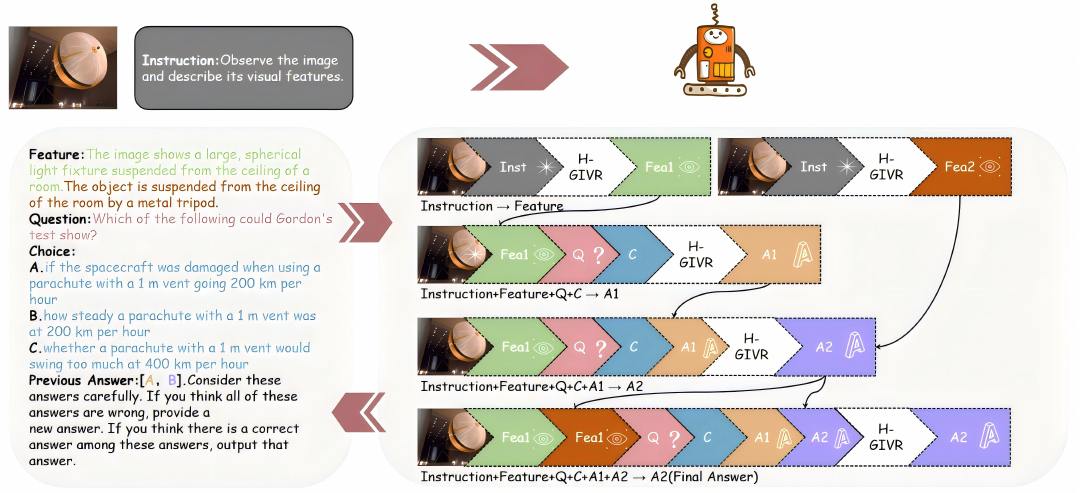}
    \caption{The detailed process by which the H-GIVR framework answers a VQA question.}
    \label{fig:intro}
\end{figure*}

Previous studies show that explicit prompts can guide models to generate reasoning steps~\cite{wei2022chain}, thereby improving performance on complex reasoning tasks. Moreover, repeated sampling combined with majority voting can enhance the stability and reliability of model outputs~\cite{wang2022self}. However, most existing self-consistency methods treat multiple generations as independent repetitions and do not fully use historical outputs as references for subsequent reasoning. Although the complexity and diversity of prompts substantially affect performance~\cite{fu2022complexity}, this independent-repetition strategy finds it difficult to correct dynamic error across iterations and thus limits further improvements in the models’ reasoning abilities.

A key advantage of human reasoning does not lie in mechanical repetition, but in refining conclusions through iterative verification. For example, when taking the exam,  students review their answers by combining earlier responses with the visual clues in the question: simple problems are checked quickly, while harder ones are examined repeatedly with adjusted strategies. Large language models are originally designed to imitate human conversational logic\cite{cai2024large}, yet this verification behavior is absent from current methods. Inspired by this insight, we propose the \textbf{History-Guided Iterative Visual Reasoning with Self-Correction (H-GIVR)} framework. This framework aims to equip models with a human-like verification ability, allowing them to repeatedly access visual information during reasoning and incorporate historical answers into subsequent iterations, so that the final result is obtained from iterative consistency.

Figure~\ref{fig:intro} illustrates the detailed process of how the H-GIVR framework answers a VQA question. Specifically, in visual question answering tasks, H-GIVR first guides the language model to inspect the image and generate a corresponding image description. This description is appended to the question text and serves as visual evidence for subsequent reasoning. During iterative reasoning, the answer generated in the previous round is fed back together with the question as historical context, emulating the human habit of repeatedly checking and revising answers.

When the iteration index is an even number, the framework assumes that the model has not yet formed a sufficient understanding of the visual content. It then triggers a second visual observation step, guiding the model to regenerate the image description in order to deepen and reinforce its understanding of the visual information. Once the model produces consistent answers during the iterative process, the reasoning is considered to have reached a stable state, and this answer is taken as the final output.

In summary, our contributions are as follows:
\begin{enumerate}
    \item We propose a novel framework, H-GIVR, which moves beyond traditional self-consistency methods based on ``repeated answering and majority voting''.
    \item The H-GIVR framework effectively leverages historical information and visual information to simulate human problem-solving behavior.
    \item H-GIVR uses two verification mechanisms to prompt repeated image inspection and to determine when the model has reached a stable final answer.

\end{enumerate}

\section{Related Work}
\subsection{Self-Correction of Large Language Models}
Enabling large language models (LLMs) with self-correction capability is a key approach to enhance their reliability. Current methods can be mainly divided into two categories based on intervention mechanisms: parameter adjustment during training and interactive guidance during inference~\cite{pan1}. Parameter adjustment methods, such as the reinforcement learning based by SCoRe framework~\cite{kumar2}, modify model parameters through algorithms during the training phase. They use data generated by the models themselves for training, which effectively addresses the common problems of distribution shift and behavior collapse in traditional supervised fine-tuning. However, these methods involve complex processes and high computational costs.

In contrast, inference-time guidance methods, such as RISE~\cite{qu3&6} and $S^2$R~\cite{ma4}, aim to stimulate the models’ inherent reflective potential by designing prompts. Studies have shown that even simple prompts can trigger explicit reflective behaviors in pre-trained models~\cite{Vaswani5}. Our work follows and extends the second research path, focusing on exploring how to achieve effective inference-time self-correction through structured interaction in multimodal contexts.

\subsection{Self-Consistency and Reflection}
A single round of correction is often insufficient to fix complex errors. Therefore, the mechanism of multi-round iterative self-improvement becomes a natural extension. The RISE framework~\cite{qu3&6} proposed in existing studies enables models to recursively introspect and revise outputs from all previous rounds. Furthermore, Exploratory Iteration~\cite{jiang7} highlights the importance of maintaining divergent thinking during iteration to avoid local optima. A common core of these studies is the pursuit of self-consistency. Relevant research indicates that for models to achieve effective self-improvement, they need to demonstrate key behaviors such as verification and feedback~\cite{gandhi8}. This reveals an important principle: effective self-correction is not a simple process of ``re-answering''. Instead, it is a cognitive process that involves error detection, hypothesis backtracking and logical reconstruction. Our work designs a multi-round feedback structure to systematically guide models through this cognitive process. With repeated answer correction and image descriptions as anchors for backtracking and verification, we enhance the self-consistency of cross-modal reasoning.

\subsection{Cognitive Limitations of Multimodal Models}
Although multimodal models have demonstrated outstanding performance in numerous tasks, their underlying cognitive comprehension capabilities remain questionable. Recent studies have started to systematically evaluate the fundamental flaws of large multimodal models (MLLMs). Existing research reveals that MLLMs generally lack core knowledge, such as object permanence and basic physical intuition, and this deficiency cannot be addressed by simply scaling up the models~\cite{li9}. Since models have inherent cognitive ``blind spots'', designing an external, interactive ``reflection-correction'' mechanism for them becomes an effective approach to compensate for such flaws and enhance the reliability of reasoning. It has become a trend to adopt reflection to analyze and improve model performance in the field of model adaptation~\cite{ming10}.

\subsection{Interaction and Alignment}
``How to achieve accurate cross-modal alignment'' is the core challenge in vision-language tasks. Most existing studies focus on modal fusion and representation alignment during the training phase. Currently, research on extending self-correction mechanisms to vision-language multimodal scenarios is still in its early stages, and most existing works follow the path of parametric adjustment. For instance, some studies improve the self-consistency of visual localization via fine-tuning~\cite{he11,sun12}, while others enhance the model’s robustness against perturbations~\cite{ma13}. These studies have demonstrated that pursuing consistency in multimodal contexts is a feasible and important research direction.

However, these methods do not provide models with the ability to dynamically evaluate and correct their own cross-modal understanding during the inference phase. Research on multimodal self-correction frameworks that do not rely on complex training and are fully implemented through inference-time interaction remains scarce. Our work complements the above-mentioned studies: instead of trying to train a "perfect" cross-modal alignment model at one time, we acknowledge that the existing models have inherent alignment errors. We then design a lightweight, inference-time interactive framework that enables models to independently reflect on and correct these errors.

\begin{figure*}[!t]
    \centering
    \setlength{\tabcolsep}{1pt}
    \includegraphics[width=1\textwidth]{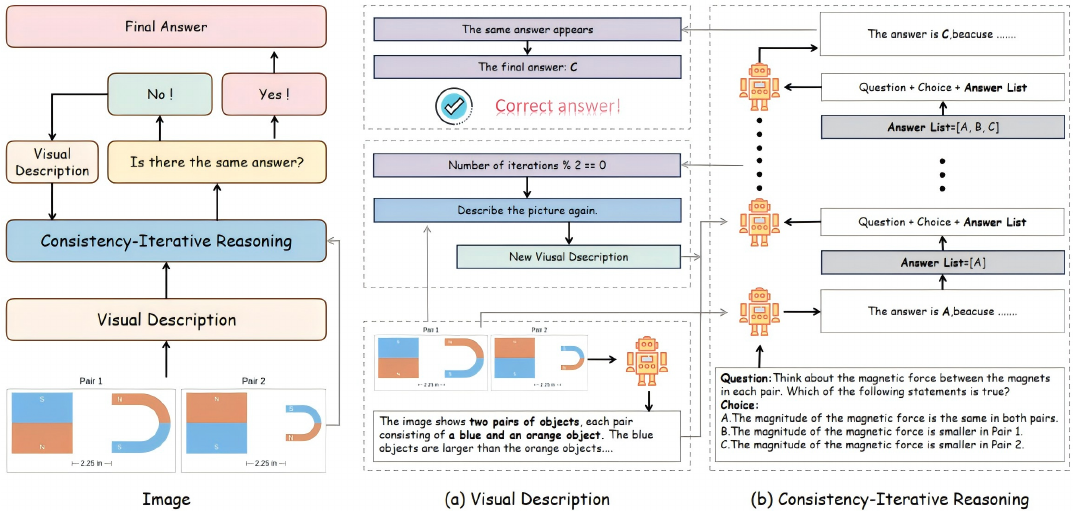}
    \caption{The illustration of H-GIVR framework. \textbf{Visual Description} and \textbf{Consistency-Iterative Reasoning} serve as the two core components, which aim to deepen the model’s understanding of images and to simulate human review behavior during the answering process; \colorbox[RGB]{218,207,223}{Image Re-observation Mechanism} and \colorbox[RGB]{218,207,223}{Answer Confirmation Mechanism} act as two key decision mechanisms, which are designed to reduce the model’s misunderstanding of visual information and to determine the final answer.}
    \label{fig:method}
\end{figure*}

\section{Methodology}
Figure~\ref{fig:method} illustrates the overall workflow of H-GIVR. The framework consists of two core components: \textbf{(i) Visual Description}, which continuously generates image descriptions throughout the entire answering process; and \textbf{(ii) Consistency-Iterative Reasoning}, where the model’s answer from the previous round is appended to the question in the next round as a reference for further reasoning. In addition, H-GIVR includes two key judgment mechanisms: \textbf{(i) Image Re-observation Mechanism}, which is activated when the iteration count is an even number, prompting the model to regenerate the image description to reduce the impact of memory loss on answer correctness; and \textbf{(ii) Answer Confirmation Mechanism}, which stops the iteration and outputs the result as the final answer once two identical answers appear during the reasoning process.

\subsection{Problem Definition}
Let a visual question answering (VQA) dataset be denoted as $D = \{(x_i,v_i,y_i)\}_{i=1}^N$, where $x_i$ represents the question, $v_i$ represents the associated image, $y_i$ represents the ground truth, and $N$ denotes the total number of test samples. For the given inputs $x_i$ and $v_i$, the image description generated by the model is defined as $f_k(v_i)$, and the answer generated by the model is defined as $p_k(x_i,v_i)$. After multiple iterations, these generated results are finally organized into an image description set $F(v_i)$ and an answer set $P(x_i,v_i)$:
\begin{equation}
\begin{aligned}
 F(v_i) &= \{f_1(v_i),f_2(v_i),...,f_n(v_i)\}, \\
 P(x_i,v_i) &= \{p_1(x_i,v_i),p_2(x_i,v_i),...,p_m(x_i,v_i)\}.
\end{aligned}
\label{1}
\end{equation}
If there exist two distinct indices $\alpha$ and $\beta$ ($1 \leq \alpha,\beta \leq m \ \& \ \alpha \neq \beta$) such that $p_\alpha(x_i,v_i) = p_\beta(x_i,v_i)$, then this answer is taken as the final output.

\subsection{Visual Description}

In visual question answering tasks, the visual information carried by the image is a key basis for producing correct answers. If the model fails to fully understand the image content, its reasoning process may gradually drift away from the correct direction, leading to wrong answers. To address this issue, we propose a stage-wise processing approach in which the image $v_i$ is handled separately from the question $x_i$ before reasoning. First, the model is guided to focus on the image $v_i$ and generate a corresponding visual description $f(v_i)$. Then, during the answer generation stage, the model performs joint reasoning by integrating the image $v_i$ and the description $f(v_i)$. This mechanism aims to strengthen the model’s perception and use of visual information, thereby improving its deeper understanding of image content and the accuracy of answer generation.

\subsection{Image Re-observation Mechanism}
In visual question answering tasks, humans often observe images continuously, repeatedly reviewing them to capture details that may have been missed and to correct errors in understanding. This dynamic observation process helps gradually refine the grasp of visual information during reasoning. However, due to limitations such as context length, existing models find it difficult to access image information multiple times within a single reasoning process, which restricts their ability to refer back to the image for self-correction.

To address this gap, we introduce a structured image re-observation mechanism into the model’s iterative reasoning process. Specifically, when the iteration count reaches an even number, the model independently reprocesses the original image and produces an updated visual description. This design is intended to simulate the human behavior of periodically looking back at the image to deepen understanding during problem solving. Through this periodic reinforcement of visual information, the model can maintain awareness of key image content across multiple reasoning steps, reduce errors caused by memory loss or insufficient initial understanding, and thereby improve the robustness and accuracy of visual reasoning.

\subsection{Consistency-Iterative Reasoning}
When reviewing answers to a question, humans do not simply repeat previous responses in a mechanical way. Instead, they examine existing answers with clear intent and purpose. To simulate this cognitive process, we move beyond traditional self-consistency methods based on repeated sampling and propose an iterative consistency reasoning component. Specifically, the H-GIVR framework maintains a dynamically updated answer list, which is appended to the question text as part of the context. After each round of reasoning, the newly generated answer is added to this list:
\begin{equation}
\mathcal{A}nswer\ List = \{a_1,a_2,...,a_n\} \label{2}
\end{equation}
where $a_i$ denotes a historical answer and $n$ represents the current iteration step. The model is then instructed to carefully consider all historical answers in the list during subsequent reasoning. This design explicitly equips the model with the ability to reflect on and verify its own outputs, making it closer to the deliberate answer-checking behavior observed in human problem solving.

At the same time, as shown in Table~\ref{tab:false}, incorporating incorrect answers as references in subsequent reasoning not only does not mislead the model, but can also encourage convergence toward the correct answer. \textcolor[RGB]{180,0,0}{\textbf{!!!!}} \textbf{In real-world applications, since it is usually not possible to determine in advance whether an answer is incorrect, this setting is mainly used for methodological validation and is not directly intended for practical deployment.}

\begin{table}[H]  
\centering
\caption{\textbf{Whether incorrect historical answers can mislead the model’s subsequent reasoning.}Standard denotes directly invoking the model, False indicates using incorrect answers as references for subsequent reasoning, and H-GIVR refers to our proposed framework.}
\label{tab:false}
\begin{tabular}{c|ccc}
\toprule
Method & Standard & False & H-GIVR \\
\midrule
Accuracy & 38.08 & 83.33 & 78.90 \\
\bottomrule
\end{tabular}
\end{table}

\subsection{Answer Confirmation Mechanism}
During the iterative process, when two identical answers appear, H-GIVR treats this answer as the final output. In addition, experiments show that the model may produce ``None'' when dealing with difficult questions. To simulate time constraints in exam settings and to improve computational efficiency, we set the maximum number of iterations to 10. According to experimental statistics, the average number of iterations required by the H-GIVR framework is at most around 4, which is well below the preset upper limit of 10. Therefore, setting the maximum number of iterations to 10 does not have a noticeable impact on the actual efficiency or performance of the framework. This setting provides sufficient room for iterative reasoning while avoiding unnecessary computational cost. Moreover, if the model generates several consecutive ``None'' outputs and then produces a valid answer for the first time, this answer is directly accepted as the final result. The pseudocode of the H-GIVR framework is shown in Algorithm~\ref{alg:vmcir}.

\begin{algorithm}[tb]
  \caption{H-GIVR}
  \label{alg:vmcir}
  \begin{algorithmic}
    \STATE {\bfseries Input:} question $x_i$, image $v_i$
    \STATE {\bfseries Initialize:} $AnswerList \leftarrow \emptyset$, $FeatureSet \leftarrow \emptyset$, $iteration \leftarrow 0$, $max\_iterations \leftarrow 10$
    \STATE Visual Description: $feature \leftarrow \text{MLLM}(v_i)$
    \STATE $FeatureSet \leftarrow FeatureSet \cup \{feature\}$
    \WHILE{$iteration < max\_iterations$}
        \STATE $iteration \leftarrow iteration + 1$
        \STATE $prompt \leftarrow x_i + FeatureSet$
        \IF{$AnswerList \neq \emptyset$}
            \STATE $prompt \leftarrow prompt + AnswerList$
        \ENDIF
        \STATE $current\_answer \leftarrow \text{MLLM}(prompt)$
        \IF{$current\_answer \in AnswerList$}
            \STATE $A_{final} \leftarrow current\_answer$
            \STATE \textbf{break}
        \ENDIF
        \STATE $AnswerList \leftarrow AnswerList \cup \{current\_answer\}$
        \IF{$iteration \mod 2 = 0$}
            \STATE  $new\_feature \leftarrow \text{MLLM}(v_i)$
            \STATE $FeatureSet \leftarrow FeatureSet \cup \{new\_feature\}$
        \ENDIF
    \ENDWHILE
    \IF{$iteration \geq max\_iterations$}
        \STATE $A_{final} \leftarrow AnswerList[-1]$ 
        \STATE{Return the latest answer}
    \ENDIF
    \STATE {\bfseries RETURN:} $A_{final}$
  \end{algorithmic}
\end{algorithm}

\begin{table*}[t]  
\centering
\caption{Overall results (\%) across five datasets, three models, and three application settings. For each setting, the best result is highlighted in bold, and the improvement gains are shown in \textcolor[RGB]{252,117,123}{pink}. The $\mathbf{\checkmark}$ symbol indicates that H-GIVR is used, while the $\mathbf{\times}$ symbol indicates that it is not used.}
\label{tab:main_results}
\small  
\begin{adjustbox}{width=\textwidth}
\renewcommand{\arraystretch}{1.3}
\begin{tabular}{c c c c c c c c c}
\hline
\multirow{2}{*}{Model} & \multirow{2}{*}{Prompt} & \multirow{2}{*}{H-GIVR} & \multicolumn{5}{c}{Dataset} & \multirow{2}{*}{Average} \\
\cline{4-8}
& & & ScienceQA & A-OKVQA & OK-VQA & VQAv2 & TextVQA &  \\
\cmidrule(r){1-1} \cmidrule(lr){2-2} \cmidrule(lr){3-3} \cmidrule(lr){4-8} \cmidrule(l){9-9}
\multirow{9}{*}{Llama3.2-vision:11b} & \multirow{3}{*}{Standard} & $\mathbf{\times}$ & 38.08 & 38.34 & 28.74 & 56.94 & 46.16 & 41.65\\

& & $\mathbf{\checkmark}$ & \textbf{78.90} & \textbf{76.42} & \textbf{48.62} & \textbf{64.99} & \textbf{57.48} & \textbf{65.28} \\
& & & \textcolor[RGB]{252,117,123}{$\uparrow 40.82$} & \textcolor[RGB]{252,117,123}{$\uparrow 38.08$} & \textcolor[RGB]{252,117,123}{$\uparrow 19.88$} & \textcolor[RGB]{252,117,123}{$\uparrow 8.05$} & \textcolor[RGB]{252,117,123}{$\uparrow 11.32$} & \textcolor[RGB]{252,117,123}{$\uparrow 23.63$} \\
\cline{2-9}
& \multirow{3}{*}{Simple CoT} & $\mathbf{\times}$ & 40.23 & 38.42 & 28.86 & 60.46 & 50.16 & 43.63 \\

& & $\mathbf{\checkmark}$ & \textbf{75.10} & \textbf{78.52} & \textbf{47.99} & \textbf{65.26} & \textbf{59.12} & \textbf{65.20} \\
& & & \textcolor[RGB]{252,117,123}{$\uparrow 34.87$} & \textcolor[RGB]{252,117,123}{$\uparrow 40.10$} & \textcolor[RGB]{252,117,123}{$\uparrow 19.13$} & \textcolor[RGB]{252,117,123}{$\uparrow 4.80$} & \textcolor[RGB]{252,117,123}{$\uparrow 8.96$} & \textcolor[RGB]{252,117,123}{$\uparrow 21.57$} \\
\cline{2-9}
& \multirow{3}{*}{Complex CoT} & $\mathbf{\times}$ & 64.47 & 40.34 & 40.11 & 62.78 & 54.56 & 52.45 \\

& & $\mathbf{\checkmark}$ & \textbf{78.14} & \textbf{68.56} & \textbf{42.49} & \textbf{64.56} & \textbf{58.44} & \textbf{62.44} \\
& & & \textcolor[RGB]{252,117,123}{$\uparrow 13.67$} & \textcolor[RGB]{252,117,123}{$\uparrow 28.22$} & \textcolor[RGB]{252,117,123}{$\uparrow 2.38$} & \textcolor[RGB]{252,117,123}{$\uparrow 1.78$} & \textcolor[RGB]{252,117,123}{$\uparrow 3.88$} & \textcolor[RGB]{252,117,123}{$\uparrow 9.99$} \\
\hline
\multirow{9}{*}{Qwen2.5vl:7b} & \multirow{3}{*}{Standard} & $\mathbf{\times}$ & 40.16 & 42.97 & 23.50 & 58.52 & 35.10 & 40.05\\

& & $\mathbf{\checkmark}$ & \textbf{85.12} & \textbf{81.83} & \textbf{55.43} & \textbf{73.22} & \textbf{68.18} &  \textbf{72.76}\\
& & & \textcolor[RGB]{252,117,123}{$\uparrow 44.96$}  & \textcolor[RGB]{252,117,123}{$\uparrow 38.86$} & \textcolor[RGB]{252,117,123}{$\uparrow 31.93$} & \textcolor[RGB]{252,117,123}{$\uparrow 14.70$} & \textcolor[RGB]{252,117,123}{$\uparrow 33.08$} & \textcolor[RGB]{252,117,123}{$\uparrow 32.71$} \\
\cline{2-9}
& \multirow{3}{*}{Simple CoT} & $\mathbf{\times}$ & 40.75 & 42.70 & 31.20 & 38.31 & 50.32 & 40.66 \\

& & $\mathbf{\checkmark}$ & \textbf{85.03} & \textbf{81.57} & \textbf{55.35} & \textbf{73.19} & \textbf{65.96} & \textbf{72.22} \\
& & & \textcolor[RGB]{252,117,123}{$\uparrow 44.28$} & \textcolor[RGB]{252,117,123}{$\uparrow 38.87$} & \textcolor[RGB]{252,117,123}{$\uparrow 24.15$} & \textcolor[RGB]{252,117,123}{$\uparrow 34.88$} & \textcolor[RGB]{252,117,123}{$\uparrow 15.64$} & \textcolor[RGB]{252,117,123}{$\uparrow 31.56$} \\
\cline{2-9}
& \multirow{3}{*}{Complex CoT} & $\mathbf{\times}$ & 73.83 & 39.56 & 40.24 & 58.96 & 61.70 & 54.86 \\

& & $\mathbf{\checkmark}$ & \textbf{84.93} & \textbf{78.08} & \textbf{52.20} & \textbf{70.25} & \textbf{66.08} & \textbf{70.31} \\
& & & \textcolor[RGB]{252,117,123}{$\uparrow 11.10$} & \textcolor[RGB]{252,117,123}{$\uparrow 38.52$} & \textcolor[RGB]{252,117,123}{$\uparrow 11.96$} & \textcolor[RGB]{252,117,123}{$\uparrow 11.29$} & \textcolor[RGB]{252,117,123}{$\uparrow 4.38$} & \textcolor[RGB]{252,117,123}{$\uparrow 15.45$} \\
\hline
\multirow{9}{*}{Gemma3:12b} & \multirow{3}{*}{Standard} & $\mathbf{\times}$ & 40.16 & 41.57 & 33.74 & 29.19 & \textbf{56.38} & 40.21\\

& & $\mathbf{\checkmark}$ & \textbf{86.14} & \textbf{83.93} & \textbf{50.09} & \textbf{65.89} & \textbf{55.76} & \textbf{68.36} \\
& & & \textcolor[RGB]{252,117,123}{$\uparrow 45.98$}  & \textcolor[RGB]{252,117,123}{$\uparrow 42.36$} & \textcolor[RGB]{252,117,123}{$\uparrow 16.35$} & \textcolor[RGB]{252,117,123}{$\uparrow 36.70$} & \textcolor[RGB]{180,180,180}{$\downarrow 0.62$} & \textcolor[RGB]{252,117,123}{$\uparrow 28.15$} \\
\cline{2-9}
& \multirow{3}{*}{Simple CoT} & $\mathbf{\times}$ & 40.75 & 44.19 & 35.00 & 38.17 & 54.34 & 42.49 \\

& & $\mathbf{\checkmark}$ & \textbf{86.39} & \textbf{83.49} & \textbf{51.31} & \textbf{66.27} & \textbf{55.72} & \textbf{68.64} \\
& & & \textcolor[RGB]{252,117,123}{$\uparrow 45.64$} & \textcolor[RGB]{252,117,123}{$\uparrow 39.30$} & \textcolor[RGB]{252,117,123}{$\uparrow 16.31$} & \textcolor[RGB]{252,117,123}{$\uparrow 28.10$} & \textcolor[RGB]{252,117,123}{$\uparrow 1.38$} & \textcolor[RGB]{252,117,123}{$\uparrow 26.15$} \\
\cline{2-9}
& \multirow{3}{*}{Complex CoT} & $\mathbf{\times}$ & 61.90 & 41.92 & 45.20 & 59.03 & 41.88 & 49.99 \\

& & $\mathbf{\checkmark}$ & \textbf{84.58} & \textbf{81.92} & \textbf{50.56} & \textbf{66.37} & \textbf{55.22} & \textbf{67.73} \\
& & & \textcolor[RGB]{252,117,123}{$\uparrow 22.69$} & \textcolor[RGB]{252,117,123}{$\uparrow 40.00$} & \textcolor[RGB]{252,117,123}{$\uparrow 5.36$} & \textcolor[RGB]{252,117,123}{$\uparrow 7.34$} & \textcolor[RGB]{252,117,123}{$\uparrow 13.34$} & \textcolor[RGB]{252,117,123}{$\uparrow 17.74$} \\
\hline

\end{tabular}
\end{adjustbox}
\end{table*}

\section{Experiments}
In this section, we conduct extensive experiments to evaluate the effectiveness and computational efficiency of H-GIVR. Our experiments mainly focus on addressing the following key questions:
\begin{enumerate}
    \item [\textbf{Q1.}] {How does H-GIVR perform across multiple scenarios?}
    \item [\textbf{Q2.}] {Does H-GIVR require a large amount of computational resources?}
    \item [\textbf{Q3.}] {Compared with existing baseline methods, can H-GIVR improve accuracy?}
    \item [\textbf{Q4.}] {Both Visual Description and Consistency-Iterative Reasoning can effectively improve accuracy?}
    \item [\textbf{Q5.}] {Is H-GIVR still effective in specific domains?}
\end{enumerate}

\subsection{Datasets and Metrics}
Following the standard evaluation settings in MLLMs reasoning research, we conduct experiments on five popular benchmarks: ScienceQA~\cite{lu2022learn}, A-OKVQA~\cite{schwenk2022okvqa}, OK-VQA~\cite{marino2019ok}, VQAv2~\cite{goyal2017making} and TextVQA~\cite{singh2019towards}. Among the datasets, ScienceQA and A-OKVQA are multiple-choice datasets, while OK-VQA, VQAv2, and TextVQA are open-ended answer datasets. We follow the practice~\cite{liu2024improved,li2024llava,bai2023qwen} of using accuracy as the metric. We conduct experiments using three models that each have over one million downloads on Ollama: \texttt{Llama3.2-vision:11b}, \texttt{Qwen2.5vl:7b} and \texttt{Gemma3:12b}.

\begin{table*}[t]  
\centering
\caption{Overall results (\%) of two models across five datasets and six methods. The best result is highlighted in bold, and the \textcolor[RGB]{252,117,123}{$\Delta Gains$} row reports the improvement of H-GIVR over the best baseline accuracy.}
\label{tab:model_performance}
\small  
\begin{adjustbox}{width=\textwidth}
\renewcommand{\arraystretch}{1.3}
\begin{tabular}{c c c c c c c c}
\hline
\multirow{2}{*}{Model} & \multirow{2}{*}{Prompt} & \multicolumn{5}{c}{Dataset} & \multirow{2}{*}{Average} \\
\cline{3-7}
& & ScienceQA & A-OKVQA & OK-VQA & VQAv2 & TextVQA &  \\
\cmidrule(r){1-1} \cmidrule(lr){2-2} \cmidrule(lr){3-7} \cmidrule(l){8-8}

\multirow{6}{*}{Llama3.2-vision:11b} & Standard & 38.08 & 38.34 & 28.74 & 56.94 & 46.16 & 41.65\\
& FS-CoT & 64.47 & 40.34 & 40.11 & 62.78 & 54.56 & 52.45 \\
& Auto-CoT & 39.07 & 59.82 & 48.13 & 63.57 & 54.40 & 53.00 \\
& Active-Pro & 40.44 & 42.62 & 42.62 & 60.12 & 56.04 & 48.37\\
& Self-Con & 50.00 & 54.41 & 43.56 & \textbf{66.04} & 54.85 & 53.77 \\
& \textbf{Ours} 
& \textbf{78.90} 
& \textbf{76.42} 
& \textbf{48.62} 
& 64.99 
& \textbf{57.48} 
& \textbf{65.28} \\
& $\textcolor[RGB]{252,117,123}{\Delta Gains}$ 
& \textcolor[RGB]{252,117,123}{$\uparrow 14.43$}
& \textcolor[RGB]{252,117,123}{$\uparrow 16.60$}
& \textcolor[RGB]{252,117,123}{$\uparrow 0.49$}
& \textcolor[RGB]{180,180,180}{$\downarrow 1.05$}
& \textcolor[RGB]{252,117,123}{$\uparrow 1.44$}
& \textcolor[RGB]{252,117,123}{$\uparrow 11.51$}\\
\hline

\multirow{6}{*}{Qwen2.5vl:7b} & Standard & 40.16 & 42.97 & 23.50 & 58.52 & 35.10 & 40.05\\
& FS-CoT & 73.83 & 39.56 & 40.24 & 58.96 & 61.70 & 54.86 \\
& Auto-CoT & 74.09 & 69.78 & 39.46 & 66.79 & 63.64 & 62.75 \\
& Active-Pro & 77.79 & 71.26 & 43.94 & 57.77 & 64.10 & 62.97 \\
& Self-Con & 74.70 & 64.45 & 54.10 & 40.43 & 43.82 & 55.50 \\
& \textbf{Ours} 
& \textbf{85.12} 
& \textbf {81.83} 
& \textbf{55.43} 
& \textbf{73.22} 
& \textbf{68.18} 
& \textbf{72.76} \\
& $\textcolor[RGB]{252,117,123}{\Delta Gains}$ 
& \textcolor[RGB]{252,117,123}{$\uparrow 7.33$}
& \textcolor[RGB]{252,117,123}{$\uparrow 10.57$}
& \textcolor[RGB]{252,117,123}{$\uparrow 1.33$}
& \textcolor[RGB]{252,117,123}{$\uparrow 6.43$}
& \textcolor[RGB]{252,117,123}{$\uparrow 4.08$}
& \textcolor[RGB]{252,117,123}{$\uparrow 9.79$}\\
\hline

\end{tabular}
\end{adjustbox}
\end{table*}

\begin{table*}[!t]
\centering
\caption{Results of ablation study ($\%$) on five benchmarks. \textbf{Visual} denotes the Visual Description module, and \textbf{Iterative} denotes the Consistency-Iterative Reasoning module. The \CheckmarkBold symbol indicates that the corresponding module is used, and \textcolor[RGB]{252,117,123}{the pink} shows the improvement compared with the Standard setting.}
\begin{tabularx}{\textwidth}{@{}l|*{2}{>{\centering\arraybackslash}>{\hsize=0.7\hsize}X}|*{5}{>{\centering\arraybackslash}X}|>{\centering\arraybackslash}>{\hsize=0.8\hsize}X@{}}
\toprule
\multirow{2}{*}{} & \multicolumn{2}{c|}{Module} & \multicolumn{5}{c|}{Datasets} & \multirow{2}{*}{Avg.} \\
\cmidrule(lr){2-3} \cmidrule(lr){4-8}
&  Visual &  Iterative &  ScienceQA &  A-OKVQA &  OK-VQA &  VQAv2 &  TextVQA \\
\midrule

& & & \multicolumn{5}{@{}c}{\textbf{Qwen2.5vl:7b}} & \\
\midrule

(a) & \multicolumn{2}{c|}{Standard} & 40.16 & 42.97 & 23.50 & 58.52 & 35.10 & 40.05 \\
\midrule
(b) & \CheckmarkBold & & 82.79 & 80.96 & 53.36 & 72.36 & 65.20 & 70.93 \\
 & & 
 & \textcolor[RGB]{252,117,123}{$\uparrow 42.63$} 
 & \textcolor[RGB]{252,117,123}{$\uparrow 37.99$} 
 & \textcolor[RGB]{252,117,123}{$\uparrow 29.86$} 
 & \textcolor[RGB]{252,117,123}{$\uparrow 13.84$} 
 & \textcolor[RGB]{252,117,123}{$\uparrow 30.10$} 
 & \textcolor[RGB]{252,117,123}{$\uparrow 30.88$} \\
(c) & & \CheckmarkBold & 83.28 & 81.79 & 54.92 & 72.45 & 68.14 & 72.11 \\
 & & 
 & \textcolor[RGB]{252,117,123}{$\uparrow 43.12$} 
 & \textcolor[RGB]{252,117,123}{$\uparrow 38.82$} 
 & \textcolor[RGB]{252,117,123}{$\uparrow 31.42$} 
 & \textcolor[RGB]{252,117,123}{$\uparrow 13.93$} 
 & \textcolor[RGB]{252,117,123}{$\uparrow 33.04$} 
 & \textcolor[RGB]{252,117,123}{$\uparrow 32.06$} \\
(d) & \CheckmarkBold & \CheckmarkBold & \textbf{85.12} & \textbf {81.83} & \textbf{55.43} & \textbf{73.22} & \textbf{68.18} &  \textbf{72.76} \\
 & & 
 & \textcolor[RGB]{252,117,123}{$\uparrow 44.96$} 
 & \textcolor[RGB]{252,117,123}{$\uparrow 38.86$} 
 & \textcolor[RGB]{252,117,123}{$\uparrow 31.93$} 
 & \textcolor[RGB]{252,117,123}{$\uparrow 14.70$} 
 & \textcolor[RGB]{252,117,123}{$\uparrow 33.08$} 
 & \textcolor[RGB]{252,117,123}{$\uparrow 32.71$} \\

\bottomrule
\end{tabularx}
\label{tab:ablation}
\end{table*}

\subsection{Experimental results}
\paragraph{\textbf{For Q1: }How does H-GIVR perform across multiple scenarios?}
We conduct experiments under three typical application settings for multimodal large language models: Standard, which directly invokes the model; Simple CoT, which uses a simple ``Let’s think step by step'' prompt; and Complex CoT, which applies few-shot prompting. The experimental results are reported in Table~\ref{tab:main_results}.

The results show that H-GIVR achieves significant performance improvements across all five datasets, three models, and three application settings, with particularly strong gains on multiple-choice datasets. For the Llama3.2-vision, Qwen2.5vl, and Gemma3 models, performance improvements exceed 50\% in both the Standard and Simple CoT settings, and the minimum improvement in the Complex CoT setting still reaches 19\%. On the more challenging open-ended datasets, H-GIVR also provides clear benefits, with improvements ranging from 2.8\% to 69.2\%.

In addition, the results show that when extra prompt information is provided to H-GIVR, whether simple or carefully designed, the framework remains effective, but its accuracy is lower than that achieved without prompts. This indicates that H-GIVR can be used directly without any carefully crafted prompts, which helps reduce the usage barrier of multimodal large language models.

\begin{figure*}[!t]
    \centering
    \setlength{\tabcolsep}{1pt} 
    \includegraphics[width=1\textwidth]{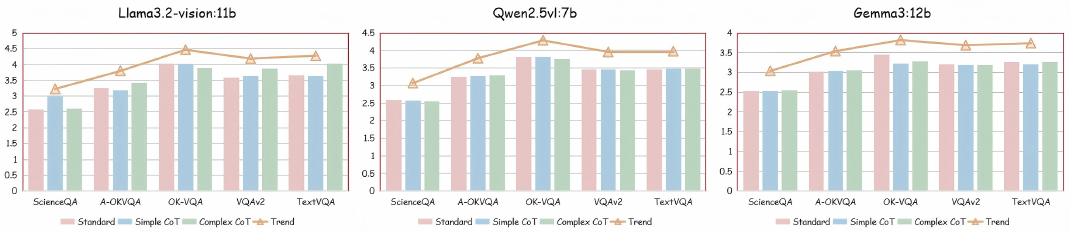}
    \caption{The average number of model calls required by H-GIVR to process each question across three models, five datasets, and three application settings. The bar chart shows the average number of model calls, while the line chart illustrates how the average number of calls varies across the five datasets.}
    \label{fig:iter}
\end{figure*}

\paragraph{\textbf{For Q2: }Does H-GIVR require a large amount of computational resources?}
We measure the average number of model calls required by H-GIVR to process each question across three models, five datasets, and three application settings, as shown in Figure~\ref{fig:iter}. The results indicate that H-GIVR requires between 2.57 and 4.04 calls per question, suggesting that H-GIVR is computationally efficient and has good cost control.

We also observe that, compared with multiple-choice questions, open-ended questions usually require more model calls. This indicates that such questions involve higher decision uncertainty and make it more difficult for the model to converge to a clear and stable answer. Since open-ended tasks allow a wider range of valid responses, the model often needs additional reasoning steps to refine its output and reach a consistent conclusion. In addition, compared with the Standard setting without prompts, the Simple CoT and Complex CoT settings with prompts generally require fewer calls, but their accuracy is often lower than that of the Standard setting. This suggests that when explicit prompts are introduced, the model may be more likely to remain fixed on previously generated incorrect answers, rather than continuing deeper reasoning and correction.

\begin{figure*}[!t]
    \centering
    \setlength{\tabcolsep}{1pt}
    \includegraphics[width=1\textwidth]{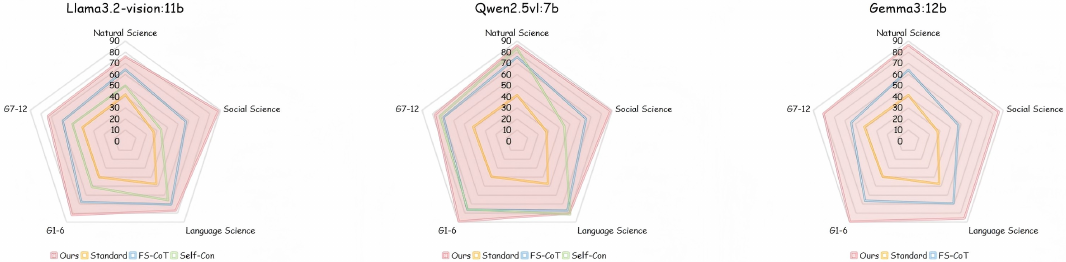}
    \caption{Performance differences between H-GIVR and other baseline methods across fine-grained domains.}
    \label{fig:field}
\end{figure*}

\paragraph{\textbf{For Q3: }Compared with existing baseline methods, can H-GIVR improve accuracy?}
We select several classic prompting methods from unimodal large language models as baselines, including few-shot chain-of-thought prompting (FS CoT), Auto-CoT~\cite{zhang2022automatic}, Active Prompt~\cite{diao2024active}, and self-consistency~\cite{wang2022self}. As shown in Table~\ref{tab:model_performance}, across five datasets, the Llama3.2-vision and Qwen2.5vl models achieve better performance in most cases when using H-GIVR, compared with other baseline methods. The performance gains range from 0.49 to 16.60.

Although Self-Consistency performs slightly better in a few cases (for example, Llama3.2-vision on the VQAv2 dataset), its number of model calls is fixed at 5, whereas the average number of calls required by H-GIVR is only 3.58. Therefore, H-GIVR maintains competitive performance while offering a clear advantage in computational efficiency.

\paragraph{\textbf{For Q4: }Both Visual Description and Consistency-Iterative Reasoning can effectively improve accuracy?}
To verify the effectiveness of the two core components in the framework, we conduct systematic ablation experiments, and the results are shown in Figure~\ref{tab:ablation}. The experiments use the Qwen2.5vl 7B model under the Standard setting, where the model achieves an average accuracy of 40.05\% across five datasets. When only the visual description module is introduced, the average performance increases significantly to 70.93\%. When only the consistency-iterative reasoning module is used, the average accuracy further improves to 72.11\%. Finally, the complete H-GIVR, which includes both modules, achieves the best performance with an average accuracy of 72.76\%, showing stable and clear improvements across all datasets.

These results confirm the effectiveness of both the visual description module and the consistency-iterative reasoning module. They also indicate that the two components are complementary, and that their combination can jointly enhance the reasoning ability and answer accuracy of multimodal large language models.

\paragraph{\textbf{For Q5: }Is H-GIVR still effective in specific domains?}
To more systematically evaluate the performance of H-GIVR across different subject types and difficulty levels, we conduct a fine-grained analysis on the ScienceQA dataset. This dataset can be divided along two dimensions. In terms of subject categories, it covers three major areas: Language Science, Social Science, and Natural Science. In terms of question difficulty, it is grouped into two levels, G1–6 and G7–12. The experimental results are shown in Figure~\ref{fig:field}.

The results show that H-GIVR performs well in most subdomains. Only in the language science category does it slightly underperform the self-consistency method in a few cases. In other subject areas, especially in social science, H-GIVR significantly improves answer accuracy, demonstrating strong subject adaptability. In addition, H-GIVR maintains stable performance advantages across different difficulty levels. These findings indicate that H-GIVR is effective across diverse subject backgrounds and shows good generalization ability.

\section{Conclusion}

H-GIVR innovatively breaks away from the traditional self-consistency paradigm of ``repeated answering and majority voting''. Instead of mechanically repeating reasoning, it treats historical reasoning information as a key reference for subsequent decisions. At the same time, it guides the model to repeatedly observe visual content during reasoning, enabling precise correction of visual understanding errors. The framework achieves significant improvements in reasoning accuracy across VQA benchmarks, while keeping computational cost controllable and relatively low.

\section*{Impact Statement}
This paper presents the History-Guided Iterative Visual Reasoning with Self-Correction (H-GIVR) framework, whose goal is to advance the field of multimodal large language model reasoning. By reducing reliance on elaborate prompt engineering and heavy model scaling, H-GIVR lowers the deployment barriers for multimodal systems, making advanced cross-modal reasoning more accessible to non-expert users. Ethically, the framework’s lightweight design minimizes computational costs, thereby reducing the carbon footprint associated with running large-scale multimodal models. In terms of societal implications, H-GIVR can potentially enhance the reliability of visual question answering systems in real-world applications such as education and accessibility tools. There are no foreseeable negative ethical consequences, and the work aligns with the broader goal of developing efficient, user-friendly AI technologies.

\nocite{langley00}

\bibliography{example_paper}

\begin{thebibliography}{31}
\providecommand{\natexlab}[1]{#1}
\providecommand{\url}[1]{\texttt{#1}}
\expandafter\ifx\csname urlstyle\endcsname\relax
  \providecommand{\doi}[1]{doi: #1}\else
  \providecommand{\doi}{doi: \begingroup \urlstyle{rm}\Url}\fi

\bibitem[Alayrac et~al.(2022)Alayrac, Donahue, Luc, Miech, Barr, Hasson, Lenc, Mensch, Millican, Reynolds, et~al.]{alayrac2022flamingo}
Alayrac, J.-B., Donahue, J., Luc, P., Miech, A., Barr, I., Hasson, Y., Lenc, K., Mensch, A., Millican, K., Reynolds, M., et~al.
\newblock Flamingo: a visual language model for few-shot learning.
\newblock \emph{Advances in neural information processing systems}, 35:\penalty0 23716--23736, 2022.

\bibitem[Bai et~al.(2023)Bai, Bai, Chu, Cui, Dang, Deng, Fan, Ge, Han, Huang, et~al.]{bai2023qwen}
Bai, J., Bai, S., Chu, Y., Cui, Z., Dang, K., Deng, X., Fan, Y., Ge, W., Han, Y., Huang, F., et~al.
\newblock Qwen technical report.
\newblock \emph{arXiv preprint arXiv:2309.16609}, 2023.

\bibitem[Cai et~al.(2024)Cai, Duan, Haslett, Wang, and Pickering]{cai2024large}
Cai, Z., Duan, X., Haslett, D., Wang, S., and Pickering, M.
\newblock Do large language models resemble humans in language use?
\newblock In \emph{Proceedings of the workshop on cognitive modeling and computational linguistics}, pp.\  37--56, 2024.

\bibitem[Diao et~al.(2024)Diao, Wang, Lin, Pan, Liu, and Zhang]{diao2024active}
Diao, S., Wang, P., Lin, Y., Pan, R., Liu, X., and Zhang, T.
\newblock Active prompting with chain-of-thought for large language models.
\newblock In \emph{Proceedings of the 62nd Annual Meeting of the Association for Computational Linguistics (Volume 1: Long Papers)}, pp.\  1330--1350, 2024.

\bibitem[Fu et~al.(2022)Fu, Peng, Sabharwal, Clark, and Khot]{fu2022complexity}
Fu, Y., Peng, H., Sabharwal, A., Clark, P., and Khot, T.
\newblock Complexity-based prompting for multi-step reasoning.
\newblock \emph{arXiv preprint arXiv:2210.00720}, 2022.

\bibitem[Gandhi et~al.(2025)Gandhi, Chakravarthy, Singh, Lile, and Goodman]{gandhi8}
Gandhi, K., Chakravarthy, A., Singh, A., Lile, N., and Goodman, N.~D.
\newblock Cognitive behaviors that enable self-improving reasoners, or, four habits of highly effective stars.
\newblock \emph{arXiv preprint arXiv:2503.01307}, 2025.

\bibitem[Goyal et~al.(2017)Goyal, Khot, Summers-Stay, Batra, and Parikh]{goyal2017making}
Goyal, Y., Khot, T., Summers-Stay, D., Batra, D., and Parikh, D.
\newblock Making the v in vqa matter: Elevating the role of image understanding in visual question answering.
\newblock In \emph{Proceedings of the IEEE conference on computer vision and pattern recognition}, pp.\  6904--6913, 2017.

\bibitem[He et~al.(2024)He, Cascante-Bonilla, Yang, Berg, and Ordonez]{he11}
He, R., Cascante-Bonilla, P., Yang, Z., Berg, A.~C., and Ordonez, V.
\newblock Improved visual grounding through self-consistent explanations.
\newblock In \emph{Proceedings of the IEEE/CVF Conference on Computer Vision and Pattern Recognition}, pp.\  13095--13105, 2024.

\bibitem[Hoffmann et~al.(2022)Hoffmann, Borgeaud, Mensch, Buchatskaya, Cai, Rutherford, Casas, Hendricks, Welbl, Clark, et~al.]{hoffmann2022training}
Hoffmann, J., Borgeaud, S., Mensch, A., Buchatskaya, E., Cai, T., Rutherford, E., Casas, D. d.~L., Hendricks, L.~A., Welbl, J., Clark, A., et~al.
\newblock Training compute-optimal large language models.
\newblock \emph{arXiv preprint arXiv:2203.15556}, 2022.

\bibitem[Jiang et~al.(2025)Jiang, Lupu, and Bachrach]{jiang7}
Jiang, M., Lupu, A., and Bachrach, Y.
\newblock Bootstrapping task spaces for self-improvement.
\newblock \emph{arXiv preprint arXiv:2509.04575}, 2025.

\bibitem[Kaplan et~al.(2020)Kaplan, McCandlish, Henighan, Brown, Chess, Child, Gray, Radford, Wu, and Amodei]{kaplan2020scaling}
Kaplan, J., McCandlish, S., Henighan, T., Brown, T.~B., Chess, B., Child, R., Gray, S., Radford, A., Wu, J., and Amodei, D.
\newblock Scaling laws for neural language models.
\newblock \emph{arXiv preprint arXiv:2001.08361}, 2020.

\bibitem[Kumar et~al.(2024)Kumar, Zhuang, Agarwal, Su, Co-Reyes, Singh, Baumli, Iqbal, Bishop, Roelofs, Zhang, McKinney, Shrivastava, Paduraru, Tucker, Precup, Behbahani, and Faust]{kumar2}
Kumar, A., Zhuang, V., Agarwal, R., Su, Y., Co-Reyes, J.~D., Singh, A., Baumli, K., Iqbal, S., Bishop, C., Roelofs, R., Zhang, L.~M., McKinney, K., Shrivastava, D., Paduraru, C., Tucker, G., Precup, D., Behbahani, F., and Faust, A.
\newblock Training language models to self-correct via reinforcement learning.
\newblock \emph{arXiv preprint arXiv:2409.12917}, 2024.

\bibitem[Li et~al.(2024{\natexlab{a}})Li, Zhang, Guo, Zhang, Li, Zhang, Zhang, Zhang, Li, Liu, et~al.]{li2024llava}
Li, B., Zhang, Y., Guo, D., Zhang, R., Li, F., Zhang, H., Zhang, K., Zhang, P., Li, Y., Liu, Z., et~al.
\newblock Llava-onevision: Easy visual task transfer.
\newblock \emph{arXiv preprint arXiv:2408.03326}, 2024{\natexlab{a}}.

\bibitem[Li et~al.(2022)Li, Li, Xiong, and Hoi]{li2022blip}
Li, J., Li, D., Xiong, C., and Hoi, S.
\newblock Blip: Bootstrapping language-image pre-training for unified vision-language understanding and generation.
\newblock In \emph{International conference on machine learning}, pp.\  12888--12900. PMLR, 2022.

\bibitem[Li et~al.(2024{\natexlab{b}})Li, Gao, Zhao, Wang, Sun, Lyu, Hawkins, Vasconcelos, Golan, Luo, et~al.]{li9}
Li, Y., Gao, Q., Zhao, T., Wang, B., Sun, H., Lyu, H., Hawkins, R.~D., Vasconcelos, N., Golan, T., Luo, D., et~al.
\newblock Core knowledge deficits in multi-modal language models.
\newblock \emph{arXiv preprint arXiv:2410.10855}, 2024{\natexlab{b}}.

\bibitem[Liu et~al.(2023)Liu, Li, Wu, and Lee]{liu2023visual}
Liu, H., Li, C., Wu, Q., and Lee, Y.~J.
\newblock Visual instruction tuning.
\newblock \emph{Advances in neural information processing systems}, 36:\penalty0 34892--34916, 2023.

\bibitem[Liu et~al.(2024)Liu, Li, Li, and Lee]{liu2024improved}
Liu, H., Li, C., Li, Y., and Lee, Y.~J.
\newblock Improved baselines with visual instruction tuning.
\newblock In \emph{Proceedings of the IEEE/CVF conference on computer vision and pattern recognition}, pp.\  26296--26306, 2024.

\bibitem[Lu et~al.(2022)Lu, Mishra, Xia, Qiu, Chang, Zhu, Tafjord, Clark, and Kalyan]{lu2022learn}
Lu, P., Mishra, S., Xia, T., Qiu, L., Chang, K.-W., Zhu, S.-C., Tafjord, O., Clark, P., and Kalyan, A.
\newblock Learn to explain: Multimodal reasoning via thought chains for science question answering.
\newblock \emph{Advances in Neural Information Processing Systems}, 35:\penalty0 2507--2521, 2022.

\bibitem[Ma et~al.(2025{\natexlab{a}})Ma, Wang, Liu, Liu, Chen, Zhang, Zhou, Du, and Li]{ma4}
Ma, R., Wang, P., Liu, C., Liu, X., Chen, J., Zhang, B., Zhou, X., Du, N., and Li, J.
\newblock S$^2$r: Teaching llms to self-verify and self-correct via reinforcement learning.
\newblock \emph{arXiv preprint arXiv:2502.12853}, 2025{\natexlab{a}}.

\bibitem[Ma et~al.(2025{\natexlab{b}})Ma, Xu, Fang, Zhang, and Cui]{ma13}
Ma, X., Xu, L., Fang, L., Zhang, C., and Cui, L.
\newblock Reliable cross-modal alignment via prototype iterative construction.
\newblock In \emph{Proceedings of the 33rd ACM International Conference on Multimedia}, pp.\  3847--3855, 2025{\natexlab{b}}.

\bibitem[Marino et~al.(2019)Marino, Rastegari, Farhadi, and Mottaghi]{marino2019ok}
Marino, K., Rastegari, M., Farhadi, A., and Mottaghi, R.
\newblock Ok-vqa: A visual question answering benchmark requiring external knowledge.
\newblock In \emph{Proceedings of the IEEE/cvf conference on computer vision and pattern recognition}, pp.\  3195--3204, 2019.

\bibitem[Ming \& Li(2024)Ming and Li]{ming10}
Ming, Y. and Li, Y.
\newblock Understanding retrieval-augmented task adaptation for vision-language models.
\newblock \emph{arXiv preprint arXiv:2405.01468}, 2024.

\bibitem[Pan et~al.(2024)Pan, Saxon, Xu, Nathani, Wang, and Wang]{pan1}
Pan, L., Saxon, M., Xu, W., Nathani, D., Wang, X., and Wang, W.~Y.
\newblock Automatically correcting large language models: Surveying the landscape of diverse automated correction strategies.
\newblock \emph{Transactions of the Association for Computational Linguistics}, 12:\penalty0 484--506, 2024.

\bibitem[Qu et~al.(2024)Qu, Zhang, Garg, and Kumar]{qu3&6}
Qu, Y., Zhang, T., Garg, N., and Kumar, A.
\newblock Recursive introspection: Teaching language model agents how to self-improve.
\newblock \emph{Advances in Neural Information Processing Systems}, 37:\penalty0 55249--55285, 2024.

\bibitem[Schwenk et~al.(2022)Schwenk, Khandelwal, Clark, Marino, and Mottaghi]{schwenk2022okvqa}
Schwenk, D., Khandelwal, A., Clark, C., Marino, K., and Mottaghi, R.
\newblock A-okvqa: A benchmark for visual question answering using world knowledge.
\newblock In \emph{European conference on computer vision}, pp.\  146--162. Springer, 2022.

\bibitem[Shah et~al.(2025)Shah, Rushton, Singla, Parmar, Smith, Vanjani, Vaswani, Chaluvaraju, Hojel, Ma, et~al.]{Vaswani5}
Shah, D.~J., Rushton, P., Singla, S., Parmar, M., Smith, K., Vanjani, Y., Vaswani, A., Chaluvaraju, A., Hojel, A., Ma, A., et~al.
\newblock Rethinking reflection in pre-training.
\newblock \emph{arXiv preprint arXiv:2504.04022}, 2025.

\bibitem[Singh et~al.(2019)Singh, Natarajan, Shah, Jiang, Chen, Batra, Parikh, and Rohrbach]{singh2019towards}
Singh, A., Natarajan, V., Shah, M., Jiang, Y., Chen, X., Batra, D., Parikh, D., and Rohrbach, M.
\newblock Towards vqa models that can read.
\newblock In \emph{Proceedings of the IEEE/CVF conference on computer vision and pattern recognition}, pp.\  8317--8326, 2019.

\bibitem[Sun et~al.(2025)Sun, Chen, He, Liu, Li, and Peng]{sun12}
Sun, J., Chen, K., He, X., Liu, X., Li, K., and Peng, C.
\newblock Unitrans: Unified parameter-efficient transfer learning and multimodal alignment for large multimodal foundation model.
\newblock \emph{Computers, Materials \& Continua}, 83\penalty0 (1), 2025.

\bibitem[Wang et~al.(2022)Wang, Wei, Schuurmans, Le, Chi, Narang, Chowdhery, and Zhou]{wang2022self}
Wang, X., Wei, J., Schuurmans, D., Le, Q., Chi, E., Narang, S., Chowdhery, A., and Zhou, D.
\newblock Self-consistency improves chain of thought reasoning in language models.
\newblock \emph{arXiv preprint arXiv:2203.11171}, 2022.

\bibitem[Wei et~al.(2022)Wei, Wang, Schuurmans, Bosma, Xia, Chi, Le, Zhou, et~al.]{wei2022chain}
Wei, J., Wang, X., Schuurmans, D., Bosma, M., Xia, F., Chi, E., Le, Q.~V., Zhou, D., et~al.
\newblock Chain-of-thought prompting elicits reasoning in large language models.
\newblock \emph{Advances in neural information processing systems}, 35:\penalty0 24824--24837, 2022.

\bibitem[Zhang et~al.(2022)Zhang, Zhang, Li, and Smola]{zhang2022automatic}
Zhang, Z., Zhang, A., Li, M., and Smola, A.
\newblock Automatic chain of thought prompting in large language models.
\newblock 2022.

\end{thebibliography}
\bibliographystyle{icml2026}

\newpage
\appendix
\onecolumn

\section{Interaction Cases}
\subsection{For ScienceQA, Multiple-Choice}

\captionof{table}{A ScienceQA example answered by a Llama3.2-vision:11b, without any chain-of-thought prompts. \textcolor[RGB]{252,117,123}{Pink} indicates the MLLM's answers, \textcolor[RGB]{113,168,201}{blue} highlights the reasoning logic of using historical information as a reference, and \textcolor[RGB]{96,170,102}{green} represents newly added visual descriptions.}
\label{tab:case1}
\begin{xltabular}{\linewidth}{>{\arraybackslash}X}
\toprule
\textbf{Image:} \raisebox{-\totalheight}{\includegraphics[width=0.1\linewidth]{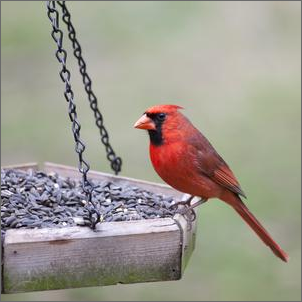}} \\
\textbf{Instruction: }Observe the image and describe its visual features. \\
\textbf{MLLM: }The image features a male Northern Cardinal perched on a wooden bird feeder, suspended from a chain. The bird's vibrant red plumage is complemented by a black mask around its eyes and a long, red tail. It is perched on the edge of the feeder, which is filled with black sunflower seeds. The background is green and blurry, suggesting that the photo was taken in a natural setting outdoors.\\
\\
\textbf{Image:} \raisebox{-\totalheight}{\includegraphics[width=0.1\linewidth]{figures/image1.png}} \\
\textbf{Question: } Identify the question that Chase's experiment can best answer.\\
\textbf{Choice: } \\
Do cardinals eat more seeds per visit from feeders containing sunflower seeds compared to feeders containing flax seeds? \\
Do cardinals visit feeders containing sunflower seeds more often than feeders containing flax seeds? \\
\textbf{Image features: }\\
The image features a male Northern Cardinal perched on a wooden bird feeder, suspended from a chain. The bird's vibrant red plumage is complemented by a black mask around its eyes and a long, red tail. It is perched on the edge of the feeder, which is filled with black sunflower seeds. The background is green and blurry, suggesting that the photo was taken in a natural setting outdoors.\\
\textbf{Instruction: }Strictly output the answer in the format ``The answer is ''. \\
\textbf{MLLM: }The answer is \textcolor[RGB]{252,117,123}{A}. \\
\\
\textbf{Image:} \raisebox{-\totalheight}{\includegraphics[width=0.1\linewidth]{figures/image1.png}} \\
\textbf{Question: } Identify the question that Chase's experiment can best answer.\\
\textbf{Choice: } \\
Do cardinals eat more seeds per visit from feeders containing sunflower seeds compared to feeders containing flax seeds? \\
Do cardinals visit feeders containing sunflower seeds more often than feeders containing flax seeds? \\
\textbf{Image features: }\\
The image features a male Northern Cardinal perched on a wooden bird feeder, suspended from a chain. The bird's vibrant red plumage is complemented by a black mask around its eyes and a long, red tail. It is perched on the edge of the feeder, which is filled with black sunflower seeds. The background is green and blurry, suggesting that the photo was taken in a natural setting outdoors.\\
\textbf{Previous answers: \textcolor[RGB]{113,168,201}{[A]. Consider these answers carefully. If you think all of these answers are wrong, provide a new answer. If you think there is a correct answer among these answers, output that answer.} } \\
\textbf{Instruction: }Strictly output the answer in the format ``The answer is ''. \\
\textbf{MLLM: }The answer is \textcolor[RGB]{252,117,123}{B}. \\
\\
\textbf{Image:} \raisebox{-\totalheight}{\includegraphics[width=0.1\linewidth]{figures/image1.png}} \\
\textbf{Instruction: }Observe the image again and describe its visual features. \\
\textbf{MLLM: }The image features a cardinal perched on a bird feeder hung from a chain. The bird's plumage is a vibrant red, with \textcolor[RGB]{96,170,102}{a black neck patch and a bright red beak}. \textcolor[RGB]{96,170,102}{It faces right, its tail feathers pointing down}. The bird feeder itself is made of wood and contains sunflower seeds, with \textcolor[RGB]{96,170,102}{a metal chain and hook used for hanging}. The surrounding environment is out of focus, but it appears to be a natural setting with greenery. \\
\\
\textbf{Image:} \raisebox{-\totalheight}{\includegraphics[width=0.1\linewidth]{figures/image1.png}} \\
\textbf{Question: } Identify the question that Chase's experiment can best answer.\\
\textbf{Choice: } \\
Do cardinals eat more seeds per visit from feeders containing sunflower seeds compared to feeders containing flax seeds? \\
Do cardinals visit feeders containing sunflower seeds more often than feeders containing flax seeds? \\
\textbf{Image features: }\\
The image features a male Northern Cardinal perched on a wooden bird feeder, suspended from a chain. The bird's vibrant red plumage is complemented by a black mask around its eyes and a long, red tail. It is perched on the edge of the feeder, which is filled with black sunflower seeds. The background is green and blurry, suggesting that the photo was taken in a natural setting outdoors.\\
The image features a cardinal perched on a bird feeder hung from a chain. The bird's plumage is a vibrant red, with \textcolor[RGB]{96,170,102}{a black neck patch and a bright red beak}. \textcolor[RGB]{96,170,102}{It faces right, its tail feathers pointing down}. The bird feeder itself is made of wood and contains sunflower seeds, with \textcolor[RGB]{96,170,102}{a metal chain and hook used for hanging}. The surrounding environment is out of focus, but it appears to be a natural setting with greenery. \\
\textbf{Previous answers: \textcolor[RGB]{113,168,201}{[A, B]. Consider these answers carefully. If you think all of these answers are wrong, provide a new answer. If you think there is a correct answer among these answers, output that answer.} } \\
\textbf{Instruction: }Strictly output the answer in the format ``The answer is ''. \\
\textbf{MLLM: }The answer is \textcolor[RGB]{252,117,123}{B}. \\

\bottomrule

\end{xltabular}

\addtocounter{table}{-1}
\captionof{table}{A ScienceQA example answered by a Llama3.2-vision:11b, without any chain-of-thought prompts. \textcolor[RGB]{252,117,123}{Pink} indicates the MLLM's answers, \textcolor[RGB]{113,168,201}{blue} highlights the reasoning logic of using historical information as a reference.}
\label{tab:case2}
\begin{xltabular}{\linewidth}{>{\arraybackslash}X}
\toprule
\textbf{Image:} \raisebox{-\totalheight}{\includegraphics[width=0.1\linewidth]{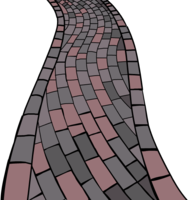}} \\
\textbf{Instruction: }Observe the image and describe its visual features. \\
\textbf{MLLM: } The provided image appears to be a 2D representation of a 3D structure, likely a road or a path, made up of numerous small blocks or bricks. The blocks seem to follow a curved path, winding through the image from top to bottom. The blocks are arranged in a pattern, with some blocks seemingly inlaid or placed at an angle, while others appear to be at a 90-degree angle to others. \\
\\
\textbf{Image:} \raisebox{-\totalheight}{\includegraphics[width=0.1\linewidth]{figures/image2.png}} \\
\textbf{Question: } Which material is this path made of?\\
\textbf{Choice: } \\
plastic \\
brick \\
\textbf{Image features: }\\
The provided image appears to be a 2D representation of a 3D structure, likely a road or a path, made up of numerous small blocks or bricks. The blocks seem to follow a curved path, winding through the image from top to bottom. The blocks are arranged in a pattern, with some blocks seemingly inlaid or placed at an angle, while others appear to be at a 90-degree angle to others.\\
\textbf{Instruction: }Strictly output the answer in the format ``The answer is ''. \\
\textbf{MLLM: }The answer is \textcolor[RGB]{252,117,123}{B}. \\
\\
\textbf{Image:} \raisebox{-\totalheight}{\includegraphics[width=0.1\linewidth]{figures/image2.png}} \\
\textbf{Question: } Which material is this path made of?\\
\textbf{Choice: } \\
plastic \\
brick \\
\textbf{Image features: }\\
The provided image appears to be a 2D representation of a 3D structure, likely a road or a path, made up of numerous small blocks or bricks. The blocks seem to follow a curved path, winding through the image from top to bottom. The blocks are arranged in a pattern, with some blocks seemingly inlaid or placed at an angle, while others appear to be at a 90-degree angle to others.\\
\textbf{Previous answers: \textcolor[RGB]{113,168,201}{[B]. Consider these answers carefully. If you think all of these answers are wrong, provide a new answer. If you think there is a correct answer among these answers, output that answer.} } \\
\textbf{Instruction: }Strictly output the answer in the format ``The answer is ''. \\
\textbf{MLLM: }The answer is \textcolor[RGB]{252,117,123}{B}. \\
\bottomrule

\end{xltabular}

\subsection{For OK-VQA, Open-Ended}

\addtocounter{table}{-1}
\captionof{table}{A OK-VQA example answered by a Qwen2.5vl:7b, without any chain-of-thought prompts. \textcolor[RGB]{252,117,123}{Pink} indicates the MLLM's answers, \textcolor[RGB]{113,168,201}{blue} highlights the reasoning logic of using historical information as a reference, and \textcolor[RGB]{96,170,102}{green} represents newly added visual descriptions.}
\label{tab:case3}
\begin{xltabular}{\linewidth}{>{\arraybackslash}X}
\toprule
\textbf{Image:} \raisebox{-\totalheight}{\includegraphics[width=0.1\linewidth]{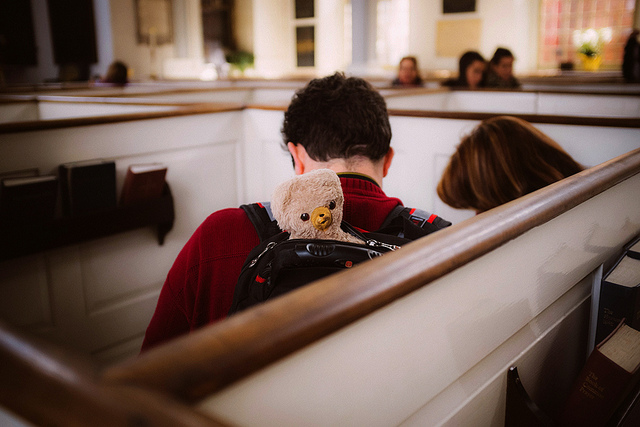}} \\
\textbf{Instruction: }Observe the image and describe its visual features. \\
\textbf{MLLM: }The image shows the back of a person sitting in a church pew, viewed from behind. \\
\\
\textbf{Image:} \raisebox{-\totalheight}{\includegraphics[width=0.1\linewidth]{figures/image3.jpg}} \\
\textbf{Question: } What toy is this?\\
\textbf{Image features: }\\
The image shows the back of a person sitting in a church pew, viewed from behind. \\
\textbf{Instruction: }Strictly output the answer in the format ``The answer is ''. \\
\textbf{MLLM: }The answer is \textcolor[RGB]{252,117,123}{teddy bear}. \\
\\
\textbf{Image:} \raisebox{-\totalheight}{\includegraphics[width=0.1\linewidth]{figures/image3.jpg}} \\
\textbf{Question: } What toy is this?\\
\textbf{Image features: }\\
The image shows the back of a person sitting in a church pew, viewed from behind. \\
\textbf{Previous answers: \textcolor[RGB]{113,168,201}{[teddy bear]. Consider these answers carefully. If you think all of these answers are wrong, provide a new answer. If you think there is a correct answer among these answers, output that answer.} } \\
\textbf{Instruction: }Strictly output the answer in the format ``The answer is ''. \\
\textbf{MLLM: }The answer is \textcolor[RGB]{252,117,123}{teddy}. \\
\\
\textbf{Image:} \raisebox{-\totalheight}{\includegraphics[width=0.1\linewidth]{figures/image3.jpg}} \\
\textbf{Instruction: }Observe the image again and describe its visual features. \\
\textbf{MLLM: }This is a cozy indoor space, likely a church chapel, bathed in soft warm light. A light-furred teddy bear peeks over the top of a black backpack, which belongs to someone in a red sweater (only their curly dark hair and back are visible). \\
\\
\textbf{Image:} \raisebox{-\totalheight}{\includegraphics[width=0.1\linewidth]{figures/image3.jpg}} \\
\textbf{Question: } What toy is this?\\
\textbf{Image features: }\\
The image shows the back of a person sitting in a church pew, viewed from behind. \\
This is a cozy indoor space, likely a church chapel, bathed in soft warm light. \textcolor[RGB]{96,170,102}{A light-furred teddy bear peeks over the top of a black backpack, which belongs to someone in a red sweater (only their curly dark hair and back are visible)}. \\
\textbf{Previous answers: \textcolor[RGB]{113,168,201}{[teddy bear, teddy]. Consider these answers carefully. If you think all of these answers are wrong, provide a new answer. If you think there is a correct answer among these answers, output that answer.} } \\
\textbf{Instruction: }Strictly output the answer in the format ``The answer is ''. \\
\textbf{MLLM: }The answer is \textcolor[RGB]{252,117,123}{teddy bear}. \\
\bottomrule

\end{xltabular}

\addtocounter{table}{-1}
\captionof{table}{A OK-VQA example answered by a Qwen2.5vl:7b, without any chain-of-thought prompts. \textcolor[RGB]{252,117,123}{Pink} indicates the MLLM's answers, \textcolor[RGB]{113,168,201}{blue} highlights the reasoning logic of using historical information as a reference.}
\label{tab:case4}
\begin{xltabular}{\linewidth}{>{\arraybackslash}X}
\toprule
\textbf{Image:} \raisebox{-\totalheight}{\includegraphics[width=0.1\linewidth]{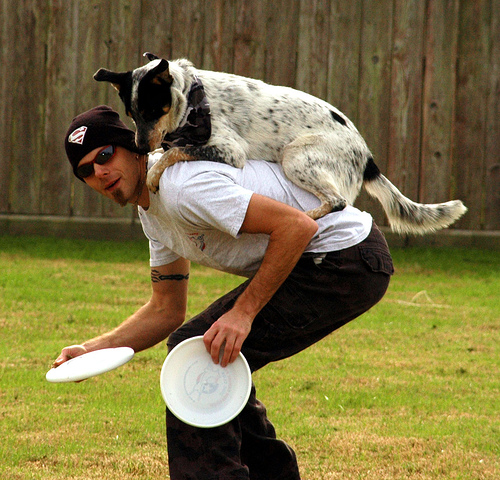}} \\
\textbf{Instruction: }Observe the image again and describe its visual features. \\
\textbf{MLLM: }The image shows a man outdoors in a grassy area, holding two white frisbees in each hand. \\
\\
\textbf{Image:} \raisebox{-\totalheight}{\includegraphics[width=0.1\linewidth]{figures/image4.jpg}} \\
\textbf{Question: }Which part of this animal would be in use of it was playing the game that is played with the items the man is holding?\\
\textbf{Image features: }\\
The image shows a man outdoors in a grassy area, holding two white frisbees in each hand. \\
\textbf{Instruction: }Strictly output the answer in the format ``The answer is ''. \\
\textbf{MLLM: }The answer is \textcolor[RGB]{252,117,123}{Mouth}. \\
\\
\textbf{Image:} \raisebox{-\totalheight}{\includegraphics[width=0.1\linewidth]{figures/image4.jpg}} \\
\textbf{Question: }Which part of this animal would be in use of it was playing the game that is played with the items the man is holding?\\
\textbf{Image features: }\\
The image shows a man outdoors in a grassy area, holding two white frisbees in each hand. \\
\textbf{Previous answers: \textcolor[RGB]{113,168,201}{[Mouth]. Consider these answers carefully. If you think all of these answers are wrong, provide a new answer. If you think there is a correct answer among these answers, output that answer.} } \\
\textbf{Instruction: }Strictly output the answer in the format ``The answer is ''. \\
\textbf{MLLM: }The answer is \textcolor[RGB]{252,117,123}{Mouth}. \\
\bottomrule

\end{xltabular}

\end{document}